\documentclass[letterpaper, 10 pt, conference]{ieeeconf}  % Comment this line out
\IEEEoverridecommandlockouts                              % This command is only
\usepackage{graphicx}
\usepackage{algorithm}
\usepackage[noend]{algpseudocode}
\title{\LARGE \bf
Automated Cinematography Motion Planning for UAVs
}

\author{Animesh Nema, Christopher Grontkowski, Derek Calzada, and Sanjuksha Nirgude}

\begin{document}

\maketitle
\thispagestyle{empty}
\pagestyle{empty}

%%%%%%%%%%%%%%%%%%%%%%%%%%%%%%%%%%%%%%%%%%%%%%%%%%%%%%%%%%%%%%%%%%%%%%%%%%%%%%%%
\begin{abstract}

This project aimed to develop an automated cinematography platform using an unmanned aerial vehicle. Quadcopters are a great platform for shooting aerial scenes but are difficult to maneuver smoothly and can require expertise to pilot. We aim to design an algorithm to enable automated cinematography of a desired object of interest.
Given the location of an object and other obstacles in the environment, the drone is able to plan its trajectory while simultaneously keeping the desired object in the video frame and avoiding obstacles.
The high maneuverability of quadcopter platforms coupled with the desire for smooth movement and stability from camera platforms means a robust motion planning algorithm must be developed which can take advantage of the quadcopter's abilities while creating motion paths which satisfy the ultimate goal of capturing aerial video. This project aims to research, develop, simulate, and test such an algorithm.

\end{abstract}

%%%%%%%%%%%%%%%%%%%%%%%%%%%%%%%%%%%%%%%%%%%%%%%%%%%%%%%%%%%%%%%%%%%%%%%%%%%%%%%%
\section{INTRODUCTION}

With applications ranging from film-making to fire-fighting to competitive racing, quadcopters have become a popular robotic platform in recent years. Driven by two pairs of counter-rotating electric motors, these devices are highly maneuverable and versatile. Due to their versatile nature and ease of accessibility of areas where human reach is expensive or dangerous, these devices are commonly used in the film making industry. These drones allow filmakers to capture shots that were previously impossible or prohibitively expensive, such as capturing wide shots of remote or hard to reach landscapes. Extremely difficult shots like follow-along close up views in sports or action-scenes. And the most aesthetically pleasing shots of fly-throughs or fly-bys of cities, oceans etc. Previously this required professional camera operators in helicopters operating expensive equipment.

Creating a perfect camera shot requires taking into account various aspects like the location and orientation of the camera with respect to the object(s), camera orientation through gimbal control, and the distance from the object(s). Hence, current use of quadcopters for aerial cinematography is mostly performed manually. Expert quadcopter operators are required for this tasks. Even with a skilled human operator it is difficult to achieve complex UAV motions, especially with co-ordinated camera gimbal movement. One of such motions is getting a 360 degree view of an object(s). In case of moving targets the difficulty increases due to the speed and unpredictability of the object. Hence, drone cinematography can be used to its fullest extent by making it autonomous. 

The research mentioned in the literature review gives a glimpse of current work on making the drone autonomous for aerial cinematography. In this paper we are focusing on cinematography of object(s) of interest in a given environment. The goal is to maneuver the quadcopter while focusing on the desired object and at the same time avoiding obstacles.

As a robotic platform, motion planning is critical for autonomous operation of a quadcopter. Their high maneuverability and under-actuation creates a non-trivial problem when creating desired motion paths. As they are generally operated at low altitudes, quadcopters face many obstacles both in the air and on the ground. This is the central topic of our project. Using the topics and methods covered in RBE550 as well other research sources, we have created an algorithm for generating motion plans for a quadcopter.

\section{PROBLEM STATEMENT}

\begin{figure}[h]
\centering
\includegraphics[width=0.3\textwidth]{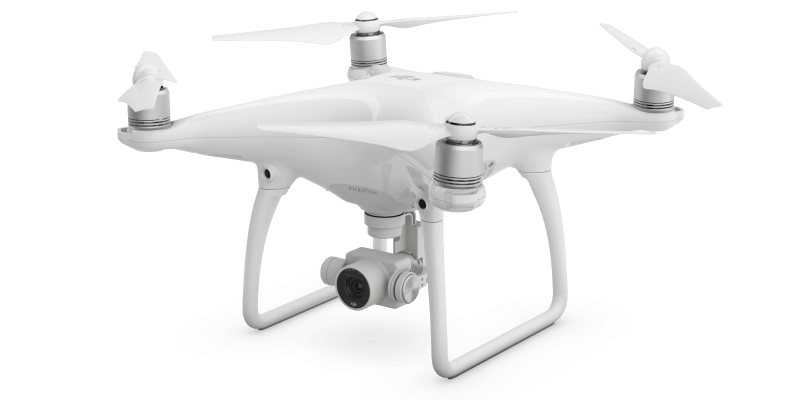}
\caption{DJI Phantom 4 quadcopter with a forward-facing, gimbal-stabilized camera.}
\label{fig:phantom4}
\end{figure}

 The principal objective of the quadcopter is to act as an aerial cinematography platform for a gimbal stabilized camera. This is one simplifying assumption for our algorithm and is based on several existing quadcopter platforms such as the DJI Phantom seen in Fig~\ref{fig:phantom4}. The gimbal compensates for small vibrations and allows the camera to tilt upwards and downwards as well as rotate in order to maintain a level horizon in the shot. Since it cannot compensate for large motions, the quadcopter still needs to maintain an orientation which points the camera in the general direction of the subject(s) to be filmed. Though this simplifies the motion paths created, the algorithm still needs to create smooth and realistic movements for the aerial platform. Without this, the footage captured by the quadcopter would be shaky and unpleasant to watch.

The movement of the quadcopter is further constrained by needing to stay within a certain range of the video's subject(s). Flying too far away can lead to the subject(s) not being visible in frame while too close means that none of the surrounding area is captured in the video. The motion path generated will also need to take into account any obstacles in the vicinity. This keeps the quadcopter from crashing. Using different interpolation and obstacle avoidance techniques, the algorithm outputs a motion path which will attempt to follow these rules while avoiding obstacles and quick changes in direction. Motions generated should avoid quick changes in direction meaning that even though a valid configuration exists in the c-space of the quadcopter, the algorithm should have a preference for those which will keep jerk low. 

The inputs to the algorithm include the quadcopter's initial position, orientation, and velocity in three dimensional space. It is also assumed that it has access to positional information of any obstacles as well as the subject(s). The outputs of our algorithm will primarily be the desired four-dimensional state of the quadcopter over time. This is composed of three positional world-coordinate states and one orientation state.

\section{LITERATURE REVIEW}
Motion planning is the central topic of this project. For our application, we need to implement a suitable motion planning algorithm for the quadcopter that can deal with the complex 3D environment. In general, the goal of the 3D path planning algorithm is to find a collision free path to or around the target as well as minimize the travel length or energy consumption. In cinematography however, the goal becomes creating motion which yields visually pleasing camera shots while actively avoiding collisions with subject(s) as well as the surroundings. 

The algorithms useful for 3D path planning can be classified into Sampling-based algorithms, node-based optimal algorithms, Mathematical model based algorithms, bio-inspired algorithms and Multi-fusion based algorithms\cite{c4}. Other algorithms include approximate path planner and geodesic path planner as explained in \cite{c5}. The approximate path planner finds a shortest path by searching and building a visibility graph. The geodesic path planner relates to wavefront-type planning, and continuously finds geodesic paths as parametric equations. Some of these are applicable to our problem statement while others find better use in other applications

\subsection{Automated Cinematography with Unmanned Aerial Vehicles}

In \cite{c1}, the authors propose and execute a motion generation algorithm for quadcopters based on the Prose Storyboard Language (PSL). This is an industry framework for describing camera shots based on framing, angle, and motion of the camera. The authors used this as inputs to automatically generate quadcopter motion paths which would achieve these shots of static and tracked moving targets. This is done by creating an artificial sphere centered around the subject(s) based on the input language commands. The commands define the size of the sphere, and the quadcopter's position on it for both the staring and ending positions. Interpolation is then performed between these two points based on the starting and ending visual properties of the shot. The system continually updates these artificial spheres and their interpolation based on the movement of the subject(s). Their algorithms were tested on both simulated and real-world quadcopters though the authors noted obstacle avoidance as a lacking feature which could be investigated in future work.

\subsection{Towards a Drone Cinematographer}

Similar to \cite{c1}, \cite{c2} proposes a method for capturing video from a quadcopter system based on industry standard shots and camera framing. The work completed by the authors allows for transitioning between multiple camera shots while still maintaining visually pleasing intermediate shots. This was again done by defining several types of shot and using an interpolation algorithm to move between the starting and ending shots. In this case, the authors performed obstacle avoidance by setting a volume within a certain distance of the subject(s) as an exclusion zone in which the drone cannot travel.

\subsection{Real-time Planning for Automated Multi-View Drone Cinematography}

The authors of \cite{c3} created and implemented a method which allows for quadcopter motion planning in an obstacle-filled and dynamic environment. This is done by using an "online receding horizon optimization formulation." This method is based around defining "virtual rails" which the authors describe as a general motion plan which is set by the user. The algorithm then attempts to follow this course while avoiding obstacles and satisfying pre-defined cinematographic objectives. They then go on to expand the capabilities of their method to allow multiple quadcopters to film simultaneously. The computed motion plans in these scenarios attempt to keep the drones from obstruction eachothers' shot while mutually avoiding collisions.

\subsection{Real-Time Motion Planning for Aerial Videography with Dynamic Obstacle Avoidance and Viewpoint Optimization }

This work \cite{c6} involves a real-time receding horizon planner based on numerical optimization that autonomously records scenes with moving targets. At the same time it also optimizes for visibility to targets and ensuring collision free trajectories. The author proposes an approach to automatic generation of quadcopter and gimbal controls in real-time while ensuring physical feasibility. The resulting appearance of the targets is specified via set-points in screen space. There are many aspects of film making that depend on different viewpoints. They have been studied and categorized forming a 'grammar' of film. In this work, the grammar of  film has been mathematically formalized for use in cost minimization algorithm. In particular, rules are formed based on how the objects specified should look on the screen. It is important to  maintain the screen position of the filmed target in order to create aesthetically pleasing footage. For our project we take an inspiration from this paper to define the desired shots. 

\subsection{Follow Moving Things in Virtual World}

\cite{c7} takes a look at tracking one or more moving targets.  By breaking down the environment into visibility-integrity regions this work is able to provide viewing positions for the camera to see the target(s).  The visibility integrity is a measure of the potential visibility of one set of points from another, taking that a step further this can essentially be used as a cost model for determining motion of the camera.  The work done in \cite{c7} also describes how to work out a projected movement of the target(s) using basic telemetry data of the target(s) and a Gaussian distribution for distance and heading change.  By looking at the the predicted location of the target(s) and the visibility-integrity roadmap, optimal camera positions are calculated from which a selected camera position is picked and a smooth path is calculated to get the camera in that position.  

%\subsection{ACT: An Autonomous Drone Cinematography System for Action Scenes}

%The authors present an automated cinematography method to capture action scenes in \cite{c8}. Their approach is based on pose estimation of the subject. The algorithm is based on first, detecting a 2D skeletal pose by using an open sourced library called Openpose and then extracting 3D skeleton points via a stereo camera. Once this has been done, the view point selection is based on the number of visible joint points (13 in this case). The trajectory planning is done on the basis of 2 criteria namely waypoint constraint and continuity constraints. i.e., whether the next waypoint exists and the trajectory is continuous. We decided to not follow the complete approach of this paper since it is focused on a motion planning based on the limb movements of a human, which is a very specific objective and cannot be replicated for other objects. However, the trajectory planning part of the paper can prove useful for our approach. 

\subsection{An interactive tool for designing quadrotor camera shots}

In \cite{c9}, authors have developed an interactive tool to create and visualize an automated videography. The authors have used Google Earth as the visualization window for any video  generated. The user needs to provide look-from and look-at points, both of which are functions of time, and changing the time-curves of these trajectories generates an automated shot. Their approach makes it easy for even novice users to capture different types of shots. The authors calculate dynamic and kinematic constraints such as gimbal joint angles, velocity, etc to account for physical constraints of the drone. This paper was helpful to us in many ways and can be used to asses reach goals such as creating complex shots and checking their feasibility in real-time.

\subsection{Steering Behaviors for Autonomous Cameras}

In \cite{c10}, the authors discuss different methods for keeping a good visual frame around the target(s) of interest as well as how to handle obstacle avoidance to keep the target(s) in frame.  The authors specifically call out the use of forces to help smooth out the camera movements.  The forces that are discussed are for target following, obstacle avoidance, camera separation, wandering and containment.  In our project only the first two are of interest, unless we added more cameras which would require the camera separation force.  These forces noted in the paper act on the camera similar to forces on a particle, this helps decide the motion of the camera so that it can smoothly be pushed and pulled around the environment and maintain some simple rules for motion.

\subsection{Intelligent Bidirectional Rapidly-exploring Random Trees}

\cite{c10} details several variations of the a Rapidly-exploring Random Tree or RRT algoritm. These include RRT, RRT*, and Bidirectional RRT*. The authors' primary focus is on developing an Intelligent Bidirectional RRT* for exploring cluttered environments. Though not directly related to aerial cinematography, the paper offers a detailed look at the related variants and go into detail explaining the various sub-functions required for a real-world implementation of the algorithms.

\section{Methodology}

\subsection{The Gazebo Environment and World Data}
The Gazebo environment that we use is custom and simplistic.  Keeping a simplistic environment allows for slow and gradual testing.  The environment consists of three obstacles, a target for the cinematography, and the quadcopter. The obstacles remain static in the environment and the target also remain static. We explored several different methods to obtain information about the environment.  With ROS and Gazebo there are a couple different methods and packages that can be used. One method that exist is using SLAM, this would be based on the lidar data coming from the quadcopter and the data could be stored in a data structure like octree.  Another method is to pull location data from ROS using services. The ROS services share all of the position and pose data of the objects. The ROS services don't provide exact geometric data about the objects so some of that data is assumed. For this project, the chosen option was using ROS services since it provided us a little bit more data than what SLAM would provide.  SLAM is only going to provide what has been seen by the quadcopter, but ROS services provides the location of all obstacles in the environment.  The location combined with some known data about the obstacles provides enough detail to do obstacle detection.  The ROS Services also allows for easy access of the quadcopter location and the location of the target which is used in the planning and motion control aspects of the application.  In a real system this would of course be replace by things like SLAM, GPS, preloaded map data, and offboard data from networked data sources. 

\subsection{Global Path Generator}

\begin{algorithm}
\caption{Global Path Generator}\label{alg:globalplan}
\begin{algorithmic}[1]
\State $path\gets GenerateArc()$ \Comment{Pass in a desired trajectory}
\State $disc\gets FindDisc(path)$ \Comment{List of discontinuities}
\State $solutions\gets None$
\For{Disc in disc}
\While{$failures < failLimit$}
\State $solution\gets RRTstar(Disc, failures)$
\If{$solution \neq None$}
\State $solutions\gets solution$
\State $break$
\Else
\State $failures\gets 1$
\EndIf
\EndWhile\label{while}
\EndFor
\State \textbf{return} $solutions$
\end{algorithmic}
\end{algorithm}

Our chosen method for path planning uses a local RRT* planner variant for obstacle avoidance while following pre-computed global paths in order to achieve a desired camera shot. The first part of the algorithm involves feeding a desired motion plan to the path planner and can be seen in Alg~\ref{alg:globalplan}. The desired motion plan is what we would consider an ideal camera motion if there were no obstacles. This can take many forms such as a simple point-to-point linear motion, or an arcing motion in which the camera is kept pointed at a target while moving around it in a circular motion. This arcing motion is one of many possible standard camera motions and one we chose to implement and focus on. The method uses circular interpolation similarly to \cite{c1} and \cite{c2} in order to compute a smooth motion path around the target. The algorithm is fed the starting point, the ending point, and the target location and outputs an arc as a series of intermediate locations which connect the two points while keeping the target in the center. This path is then fed through a collision  check in order to determine parts of the motion which would cause potential collisions and crashes. This check samples the c-space as needed and returns collision free if the quadcopter can exist at the point without colliding with an obstacle. Because of this, it is necessary to grow any obstacles in the c-space by the volume of the quadcopter plus a threshold before checking c-free. This collision check can then return what we refer to as discontinuities. Discontinuities are sections of the desired path where a collision prevents the motion from being properly executed. In a practical sense, they are start and end points from which to apply a local path planner. These points are selected along the path at a distance before and after the actual collision would occur. This makes for a smoother transition between the desired path and the locally planned path. Using this method allows us to closely follow a desired path while it is possible and only deviate when necessary. This reduces the computational load of the path planner while keeping the motion of the shot predictable by the camera operator. 

\subsection{RRT* Local Path Planner}

\begin{algorithm}
\caption{RRT* Local Path Planner}\label{alg:rrtstar}
\begin{algorithmic}[1]
\State $tree\gets startNode$ \Comment{Add start to tree}
\State $bestCost\gets \infty$
\State $bestSolution\gets None$
\While{$loopNum < maxLoops$}
\State $loopNum\gets 1$
\State $xRand\gets Sample()$ \Comment{Sandomly sample a point}
\State $xNearest\gets nearestVertex(xRand)$
\State $xNew\gets extend(xNearest)$
\State $xNew\gets bestParent(xNew)$ \Comment{Reassign parent}
\If{$NoCollision(xNew)$}
\State $tree\gets xNew$
\If{$nearGoal()$}
\If{$currentCost < bestCost$}
\State $bestCost\gets currentCost$
\State $bestSolution\gets currentSolution$
\EndIf
\EndIf
\EndIf
\EndWhile\label{while}
\State \textbf{return} $bestSolution$
\end{algorithmic}
\end{algorithm}

The local discontinuity path planner we implemented is a variant of RRT. RRTs are sample based mapping algorithms which explore the c-space of a robot. A basic RRT will have a start state and will either try to find an end state or continually expand a map until an end condition is met. This is done by randomly sampling the c-space. The closes point on the tree to the new point is then selected as the parent node and a vertex is drawn towards the random point from the parent for a set distance. If this vertex is collision-free, a new node is added at the end and the process is repeated. A variant of the RRT algorithm is known as RRT Star or RRT*. This method is different in that it will attempt to connect the new node to the node which results in the lowest cost motion from the start state. This means that over time the algorithm will trend towards the optimal path to any given node.

Our local planner is based on the RRT* algorithm with several improvements which better suit it to our application and can be seen in Alg~\ref{alg:rrtstar}. We started by studying \cite{c11} and then making changes which would better suit the algorithm to quadcopter motion planning. These include an expanding search window and an expanded RRT* neighbor search. Since our RRT* algorithm acts as a local planner, it is not very efficient in most applications to search the entirety of c-free for a path connecting discontinuities. With this in mind, we use an expanding search window and multiple iterations of the RRT* search. On the first pass, the area which RRT* can create new nodes is confined to an area slightly larger than that which would cover the discontinuity. This is more effective for areas with simple obstacle geometry such as pillars and trees. Since the environment where one is most likely to use a cinematography drone usually does not require overly complicated motion paths, assuming that an obstacle avoidance path will keep the quadcopter fairly near to the desired optimal path is a reasonable assumption. In order to account for scenarios where a larger deviation is necessary, if the RRT* search does not return a viable path after a certain number of node creations, the search window will be expanded by a set factor and the search will be re-started within this larger area. This expansion of the search area will continue through several more expansions until a viable path is discovered or the search area becomes large enough that any path through it would deviate too greatly from the desired path.

The second modification to the traditional RRT* algorithm we implemented was an expanded neighbor search. After creating a random sample point, the tree is searched for the closest existing node to that random point. The algorithm will then create a new node at a fixed radius from the closest node in the direction of the random sample. This means that the tree will reach towards the random point but will usually not directly add it as a new node. After this new node is created, traditional RRT would connect the new node with the closest one. RRT* instead connects the new node to an existing node in the immediate area which minimizes the overall cost while remaining collision free. That is, it tries to find the lowest cost path from the start state to the new node as opposed to connecting it to its immediate neighbor. We take this concept and search all the neighbors within a radius defined by the original expansion distance times an expansion factor. This has the benefit of attempting to connect nodes in long, straight lines when possible while being able to use small motions when bending around corners. In practice, this yielded smoother curves when maneuvering around obstacles and a less jagged path when the lack of obstacles allows for straight line motion.

\subsection{Visualization and Simulation}
\begin{figure}[h]
\centering
\includegraphics[width=0.3\textwidth]{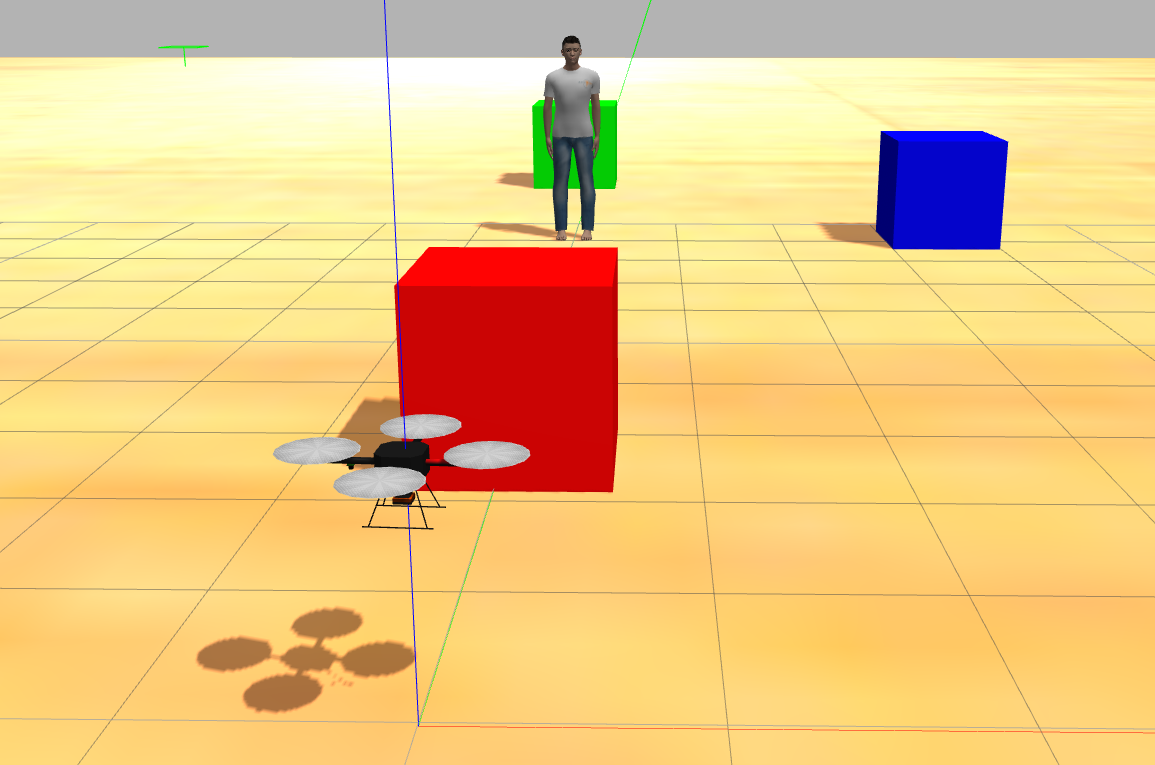}
\caption{Environment in the Gazebo Simulator.}
\label{fig:world}
\end{figure}

For this project, we used the \cite{c12}Hector quadrotor package coupled with Gazebo simulator and ROS(Robot Operating System) to simulate our approach. The hector quadrotor is a simulated robot which is built on the ROS platform. ROS topics are used as a medium of communication between the drone and its gazebo environment. We have used its Publisher/Subscriber method to publish drone velocities (linear and angluar) to the 'cmd\_vel' topic to maneuver the drone in Gazebo. We simultaneously receive the drone position and orientation in 3D space in the environment by subscribing to the '/ground\_truth\_to\_tf/pose' topic. The simulation takes place in three steps performed by a ROS node. Initially, in order to plan in the gazebo environment, the node extracts information from gazebo which includes the position, orientation and the properties of the objects in the gazebo world. This information is about the obstacles in the environment, the object of interest, here the person, and the quadrotor itself. The quadrotor is initially spawned at the center of the world. The information is then provided to the motion planner in the form of an occupancy grid. The ROS node then extracts a path from the RRT* motion planner in the form of 2D coordinates. The drone then maneuvers along the path generated by the RRT* planner as can be seen in Fig~\ref{fig:world} to reach the target location. Initially, the drone takes off due to the velocity command provided in the Z-direction to establish a desired height for the cinematographic shot. Then it takes the next point provided by RRT* planner and moves towards it following the path.

\section{Results}

\subsection{Global Path Generator}

\begin{figure}[h]
\centering
\includegraphics[width=0.45\textwidth]{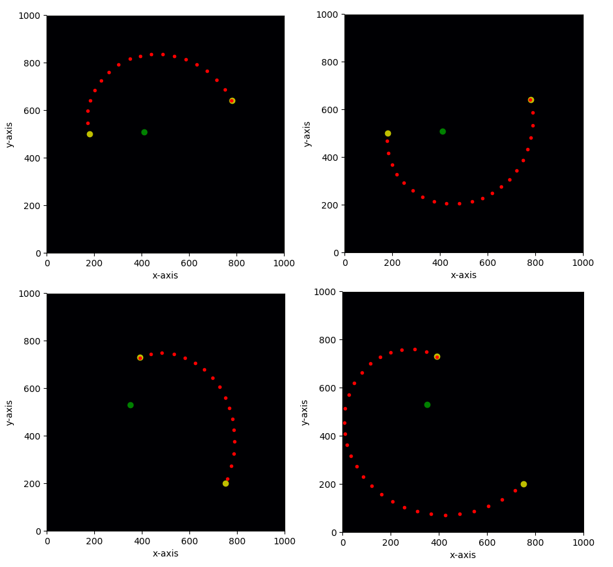}
\caption{Generation of desired quadcopter arc paths.}
\label{fig:arcpaths}
\end{figure}

Though our implementation would work with any pre-defined desired global path, we chose to focus on specific shot types for automatically generating a desired path. One of these is known as an arc and can be seen fairly often in camera based media. Using interpolation, a smooth path can be drawn between two points about a third. Fig~\ref{fig:arcpaths} demonstrates the output as a series of intermediate points which trace the arc. These arcs ignore obstacles as they only act as a guide for the final overall motion of the quadcopter. As seen, the generation of these arc works with any two start and end points as well as a target around which to revolve. With a simple boolean change, we are able to generate the arc in either a clockwise or counterclockwise direction. The arc can the be passed on to the collision detection method and RRT* planning can be performed to any section that is obstructed.

\subsection{RRT* Local Path Planner}

\begin{figure}[h]
\centering
\includegraphics[width=0.45\textwidth]{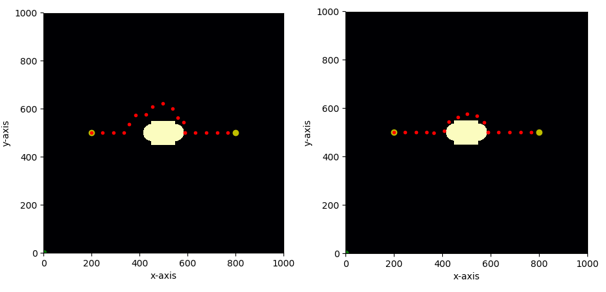}
\caption{150(Left) vs. 500(Right) RRT* loops}
\label{fig:loopnumcomparison}
\end{figure}

Our modified RRT* local planner is able to take any obstructed section of the desired motion path and generate a collision free path around it. Fig~\ref{fig:loopnumcomparison} demonstrates 150 node generations versus 500 node generations within the RRT* planner. As the number of node generations is increased, the likelihood of an optimal and smooth path around the obstacle is increased at the cost of computation time. The 150 node path took 0.126 seconds to generate while the 500 node version took 0.240 seconds. 

\begin{figure}[h]
\centering
\includegraphics[width=0.45\textwidth]{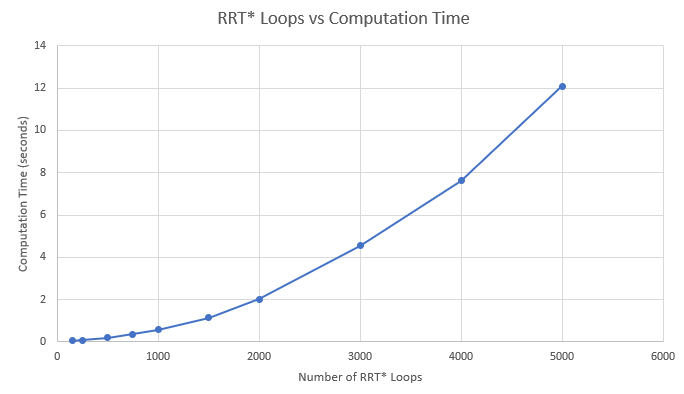}
\caption{Number of RRT* Loops vs Computation Time}
\label{fig:loopnumgraph}
\end{figure}

Fig~\ref{fig:loopnumgraph} graphs a comparison of computation time versus the number of RRT* loops. This illustrates the non-linear relationship that exists. For each scenario a balance must be struck between computation time and optimality of the graph. We found that there is no simple way to choose this balance point as it can vary with world size, obstacle shape and size, and RRT* extend distance.

\begin{figure}[h]
\centering
\includegraphics[width=0.45\textwidth]{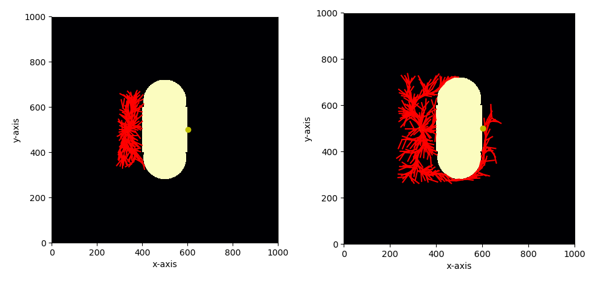}
\caption{Expansion of RRT* Graph Area}
\label{fig:expcomparison}
\end{figure}

One of the other modifications to the RRT* algorithm we incorporated was an expanding search area. Fig~\ref{fig:expcomparison} demonstrates this. During the first pass, the algorithm will search a relatively tight space around the discontinuity for a path. If this fails, the area where it can create new nodes is expanded several times until a path can be found. This method initially creates a dense node tree in an attempt to find a near optimal path to the goal. This works well for non-densely obstructed areas which is a likely environment for quadcopter cinematography. In the event that this narrow search area does not work, we included the ability to search a larger area for a path. The trade-off is in optimality of the solution with larger node areas being potentially not necessary to search. A small area increases node density but may not contain a viable solution.

\section{Future Work}
Several potential features exist which would make for a more robust motion planner but were not feasible within the time constraints of the project. Extending the algorithm to work with moving targets and obstacles is one such feature. In order to achieve this, the algorithm would need to be made more efficient in terms of implementation. This would yield quicker path results and allow for on-line motion planning. The rest of the foundation for this feature exists. The algorithm would need to be called at set time steps and provided with updated world information. The current implementation calls the planner once and then executes a path.

Another aspect to improve is the smoothness of the overall path. Though we implemented several features in an attempt to smooth the path, the random nature of RRTs means that they can sometimes produce jagged paths especially if they are not allowed to create enough nodes. A method to incorporate the direction of the desired path as it transitions into the local RRT* planner would also serve well for this application.

Finally, the last feature to implement would be a way to bias obstacle avoidance paths which minimize an obstacle blocking the drone camera's view. Different approaches exist to implement this though the easiest would be to create an obstacle shadow. That is, treat the target as a light source and treat anything behind an obstacle from this source as an obstacle even though it may exist in c-free. If a path cannot be found with the shadow enabled, the motion planner could turn it off and re-attempt a search.

\section{Conclusion}
Our team sought out to create a motion planner based on the topic learned in RBE 550 and apply them to an aerial cinematography quadcopter platform. This yeilded a global path planner which is able to follow desired trajectories based on operator defined shots as well as a local, expanding, RRT* variant for obstacle avoidance. The planner acts by taking in or generating a desired motion path for the camera. This is then searched for discontinuities where obstacles obstruct the desired path. A local RRT* path planner is applied to these sections of the path and the final, obstacle free path is output. Finally, the environment, and the quadcopter and its motion are simulated and visualized in ROS. Though some work still exist to be done as outlined in the future work section, we have created a solid foundation for an obstacle avoiding quadcopter cinematography platform.

\end{document}